\documentclass[10pt,twocolumn,letterpaper]{article}

\usepackage{iccv}
\usepackage{times}
\usepackage{epsfig}
\usepackage{graphicx}
\usepackage{amsmath}
\usepackage{amssymb}
\usepackage{multirow}
\usepackage{float} 
\usepackage{caption}
\usepackage{subcaption}
\usepackage{authblk}

\usepackage[pagebackref=true,breaklinks=true,letterpaper=true,colorlinks,bookmarks=false]{hyperref}

\iccvfinalcopy

\ificcvfinal\fi

\newcommand*\samethanks[1][\value{footnote}]{\footnotemark[#1]}

\begin{document}

\title{CCTV-Gun: Benchmarking Handgun Detection in CCTV Images}

\author[1]{Srikar Yellapragada\thanks{Authors make equal contribution to this work}}
\author[1]{\hspace{2mm} Zhenghong Li\samethanks}
\author[1]{\hspace{2mm} Kevin Bhadresh Doshi}
\author[1]{\hspace{2mm} Purva Makarand Mhasakar} 
\author[2]{\hspace{2mm} Heng Fan}
\author[3]{\hspace{2mm} Jie Wei}
\author[4]{\hspace{2mm} Erik Blasch}
\author[1]{\hspace{2mm} Bin Zhang}
\author[1]{\hspace{2mm} Haibin Ling}
\affil[1]{\normalsize Stony Brook University}
\affil[2]{\normalsize University of North Texas}
\affil[3]{City College of New York}
\affil[4]{Air Force Office of Scientific Research}
\setlength{\affilsep}{.5em}
\renewcommand\Authsep{  }
\renewcommand\Authands{  }

\maketitle

\ificcvfinal\thispagestyle{empty}\fi
\newcommand{\srikar}[1]{\textcolor[rgb]{1.00,0.00,1.00}{#1}}

\begin{abstract}
    Gun violence is a critical security problem, and it is imperative for the computer vision community to develop effective gun detection algorithms for real-world scenarios, particularly in Closed Circuit Television (CCTV) surveillance data. Despite significant progress in visual object detection, detecting guns in real-world CCTV images remains a challenging and under-explored task. Firearms, especially handguns, are typically very small in size, non-salient in appearance, and often severely occluded or indistinguishable from other small objects. Additionally, the lack of principled benchmarks and difficulty collecting relevant datasets further hinder algorithmic development. 
    In this paper, we present a meticulously crafted and annotated benchmark, called \textbf{CCTV-Gun}, which addresses the challenges of detecting handguns in real-world CCTV images. Our contribution is three-fold. Firstly, we carefully select and analyze real-world CCTV images from three datasets, manually annotate handguns and their holders, and assign each image with relevant challenge factors such as blur and occlusion. Secondly, we propose a new cross-dataset evaluation protocol in addition to the standard intra-dataset protocol, which is vital for gun detection in practical settings. Finally, we comprehensively evaluate both classical and state-of-the-art object detection algorithms, providing an in-depth analysis of their generalizing abilities. The benchmark will facilitate further research and development on this topic and ultimately enhance security. Code, annotations, and trained models are available at https://github.com/srikarym/CCTV-Gun.
\end{abstract}

\section{Introduction}
Gun violence has been a severe security problem for a long time in many countries, especially the United States \cite{lopez2018america}. Many gun-related crimes, such as robbery and shootings, occur in public places with surveillance systems. However, reliance on human supervision demands an impractical amount of vigilance, which can be expensive. Since most public surveillance systems are Closed Circuit Television (CCTV) cameras, automatic and fast detection of handguns in real-world CCTV imagery has the potential to prevent gun-related violence and/or increase interdiction response. Such a detection algorithm can potentially alert law enforcement agencies when an incident occurs. This study focuses mainly on handguns, the most commonly used type of gun in gun crimes~\cite{zawitz1995guns}. 

\begin{figure}[t]
\begin{center}
\includegraphics[width=0.49\linewidth]{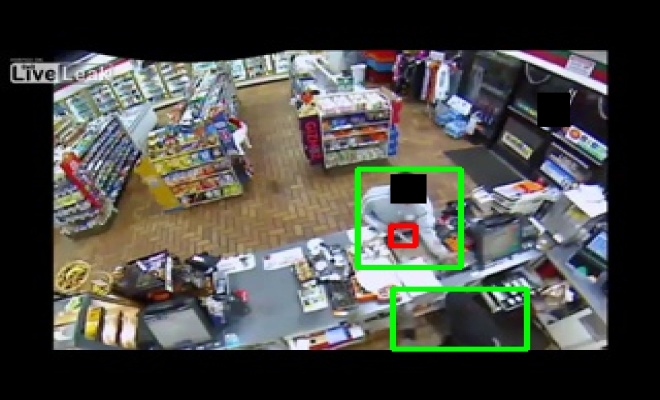} \hfill
\includegraphics[width=0.49\linewidth]{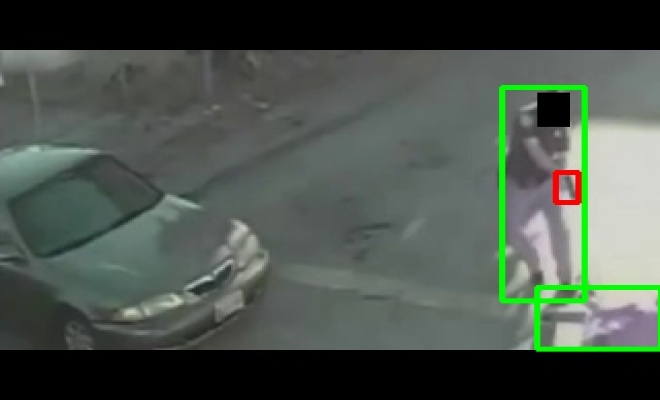}
\end{center}
  \caption{Example surveillance images involving handgun violence from the UCF-Crime dataset \cite{sultani2018real}. Handguns in low-resolution images are tiny and seriously occluded by hands, so there are no salient textures and shapes for detection. CCTV-gun annotations include both handguns and their holders. Facial regions are blocked for privacy protection.}
\label{fig:imgeg_real}
\end{figure}

In recent years, many effective deep-learning-based object detectors \cite{ren2015faster, liu2021swin, zhu2020deformable, qiao2021detectors} have been proposed. Handgun detection in real-world crime imagery is much more challenging than general object detection tasks. First, the size of handguns is too small (e.g., a few pixels) in these images. The frames in crime CCTV videos are only in 320x240 resolution, lower than current object detection datasets \cite{lin2014microsoft}. Handguns usually occupy a small area in these images, which means there are no salient texture features, and their features may be obscured in the networks. Second, the holder's hands often occlude the handguns at crime scenes. Only the barrels, which are merely in slender rectangles, can be seen in many images; see Figure \ref{fig:imgeg_real} for examples. Therefore handguns are easily misclassified with the background clutter since there are no salient shape or texture features. Third, the wide variety of camera angles and lighting conditions make detection even more difficult. 

\begin{figure*}
     \centering
     \begin{subfigure}[b]{0.33\textwidth}
         \centering
         \includegraphics[width=\textwidth]{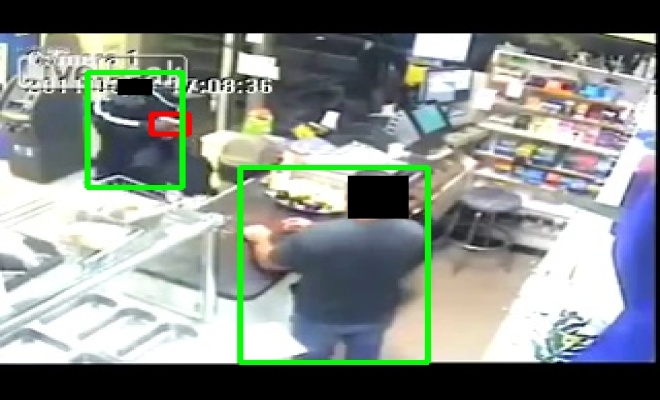}
         \caption{Occlusion (UCF) }
     \end{subfigure}
     \hfill
     \begin{subfigure}[b]{0.33\textwidth}
         \centering
         \includegraphics[width=\textwidth]{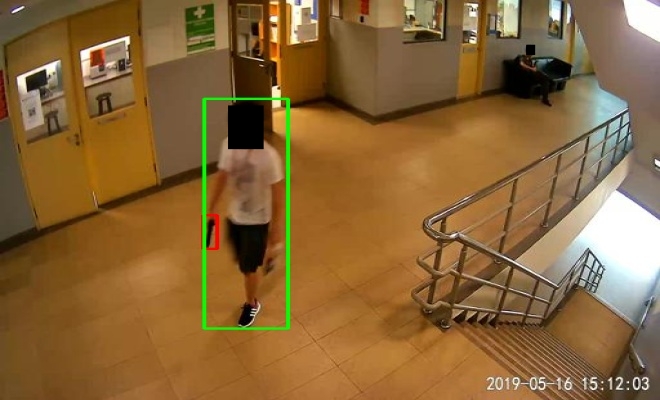}
         \caption{Blur (MGD)}
     \end{subfigure}
     \hfill
     \begin{subfigure}[b]{0.33\textwidth}
         \centering
         \includegraphics[width=\textwidth]{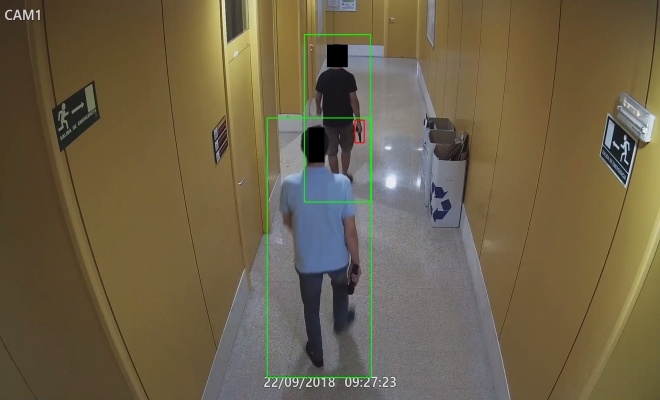}
         \caption{Similar objects (USRT)}
     \end{subfigure}
\vspace{-2mm}
\caption{Examples of challenging CCTV images for handgun detection. In ``Occlusion", handguns are partially visible. Images in ``Blur" are blurry due to motion. In ``Similar objects", people hold objects similar to handguns. From left to right, the images are originally from MGD \cite{lim2021deep}, USRT \cite{gonzalez2020real}, and UCF \cite{sultani2018real} datasets, respectively.  }
        \label{fig:challenge_gt}
\end{figure*}

Some preliminary studies of gun detection from images/videos have been conducted based on generic object detectors. Most of these studies (\eg, \cite{grega2016automated, olmos2018automatic, qi2021dataset}) focus on well-processed gun images very different from images in real crime scenes. Only minimal efforts \cite{gonzalez2020real, lim2021deep} pay some attention to real-world scenarios with CCTV images but are restricted in either data size or use of actors for simulation. Moreover, there needs to be more evaluation of state-of-the-art (SOTA) visual detection algorithms for gun detection tasks, let alone a more complicated yet critical study on their generalization capability.

To address the above challenges, this paper presents \textbf{CCTV-Gun}, a meticulously crafted and annotated benchmark for real-world handgun detection from CCTV images. Our work tackles handgun detection comprehensively in three aspects: benchmark construction, evaluation protocol, and thorough experiments. 

For \textit{benchmark construction}, we first investigate relevant real-world CCTV imagery datasets and judiciously select images from three of them: Monash Gun Dataset (MGD) \cite{lim2021deep}, US Real-time Gun detection dataset (USRT) \cite{gonzalez2020real}, and UCF Crime Scene dataset (UCF) \cite{sultani2018real}. MGD and USRT datasets contain images of enacted crime scenes, while UCF is a general-purpose action recognition dataset. We extract frames from these datasets and provide bounding box annotations of person, handgun, and handgun-holder pairs (which person holds each handgun) for all images. Moreover, for each selected image, we label it with challenge factors (\eg, blur), which helps analyze the performance of detection algorithms.

For \textit{evaluation protocol}, we propose two types of experiments: intra-dataset and cross-dataset testing. Intra-dataset is the standard evaluation technique, where a model is trained on the training split and evaluated on the test split of a given dataset. In Cross-dataset testing, we train a model on two datasets, say MGD and USRT, and test it on the entirety of the third dataset -- UCF. Cross-dataset evaluation tells us about the generalization capability of the model. We also take the model trained on two datasets (from the previous experiment), fine-tune it on the training split, and evaluate it on the test split of the third dataset. The fine-tuning evaluation tells us if models pre-trained on gun-detection datasets act as a better initialization than the COCO-pre-trained model. 

For \textit{evaluation}, we comprehensively test both classical CNN-based object detectors and state-of-the-art (SOTA) transformer-based detectors in all protocols. We also provide in-depth results analysis and insight for future directions. 

We believe this benchmark will facilitate further research on this topic and ultimately enhance security. In summary, our main contributions are as follows:
\begin{itemize}
    \vspace{-1.3mm}\item design the first carefully annotated benchmark, \textbf{CCTV-Gun}, for handgun detection in CCTV images, 
    \vspace{-1.3mm}\item develop a new cross-dataset evaluation protocol in addition to the standard intra-dataset protocol, which is vital for gun detection, and
    \vspace{-1.3mm} \item provide thorough evaluation and analysis of SOTA object detection algorithms for handgun detection.
\end{itemize}

\section{Related Work}
\subsection{Object detection}

This subsection reviews related work in generic object detection roughly in three categories. We will then summarize previous studies on gun detection in the next subsection.

\vspace{1mm}\noindent\textbf{Two Stage Methods.}
R-CNN \cite{girshick2015region} introduces the first two-stage detection algorithm. It generates region proposals using selective search, computes CNN features, and classifies them using an SVM. Fast R-CNN \cite{girshick2015fast} uses ROI Pooling, jointly learning to detect spatial locations of objects and classify them. Faster R-CNN \cite{ren2015faster} uses Region Proposal Network instead of selective search by RPN and Fast R-CNN and sharing convolution layers, making it quicker and more accurate. 

FPN \cite{lin2017feature} uses Feature Pyramids with lateral connections to obtain independent predictions at all levels and show improvements in performance by plugging in the Feature Pyramid architecture in existing methods such as Faster RCNN. Cascade RCNN \cite{cai2018cascade} observes that a single detector can only be optimal for a single IoU threshold. Cascade RCNN trains a sequence of R-CNNs, using the output of one stage to train the next. 

Recently proposed DetectoRS \cite{qiao2021detectors} uses recursive feature pyramids, which provide feedback from the top-down to bottom-up layers of an FPN. DetectoRS also introduces switchable atrous convolution that looks twice at input features with different atrous rates and combines the output.

\vspace{1mm}\noindent\textbf{One Stage Methods.}
One-stage detection models skip the region proposal phase, and detection occurs directly upon the dense sampling of locations. 
YOLOv1 \cite{redmon2016you} is the first one-stage detection algorithm. It considered detection a regression task and used an improved GoogLeNet as the backbone. It is later extended to YOLOv2~\cite{redmon2017yolo9000} by drawing inspiration from Faster R-CNN and the anchor mechanism.
A further improved version, YOLOv3~\cite{redmon2018yolov3}, uses better backbones to enhance performance. 

Another representative one-stage method is the SSD algorithm~\cite{liu2016ssd}, which focuses on improving the accuracy and recall rate. Later, the focal loss is introduced~\cite{lin2017focal} to solve the mismatch between positive and negative samples. 

\vspace{1mm}\noindent\textbf{Transformer-based Methods.}
Transformers \cite{vaswani2017attention} have been the de-facto choice of architecture in NLP. Since the advent of Vision Transformer \cite{dosovitskiy2020image}, they have achieved remarkable results in computer vision tasks such as object detection.

The first representative work in this category is DETR \cite{carion2020end}, which eliminates the intermediate steps like non-maximal suppression and anchor generation in CNN-based detectors. Instead, DETR uses a Transformer on CNN image features to predict all objects simultaneously directly. It is trained end-to-end with a set prediction loss that performs bipartite matching between predicted and ground truth objects.  

DETR takes a long time to converge compared to previous CNN-based approaches. Due to the design of the Transformer's attention module, a high-resolution feature map is required to make DETR focus on smaller objects, which increases the computation cost tremendously. Deformable DETR \cite{zhu2020deformable} solves these issues using multi-scale deformable attention modules. More recently, transformer-based detection algorithms keep pushing the front end of detection performance; some sampled studies can be found in~\cite{li2022dn,zhang2022dino}. \cite{li2022dn} feeds ground-truth boxes with noise to DETR to accelerate the training. It is also noteworthy that transformer-based backbones like Swin \cite{liu2021swin} also achieve promising performance in object detection.

\subsection{Firearm Detection}

There have been some preliminary studies on gun detection, but it remains a seriously underexplored area. A dataset from CCTV recordings with an actor is created in \cite{grega2016automated} for gun and knife detection. \cite{grega2016automated} selects positive examples for gun detection by annotating the frames in the video with a gun. They perform preprocessing with canny edge detection and PCA and classify the frame with a 2-layer neural network for the presence of a handgun, but do not detect the handgun region / bounding box. Their dataset cannot be used to train an object detection model since all the videos are shot in the same room with the same actor, thus lacking diversity.  

Internet Movie Firearms Database (IMFDB) \cite{imfdb} is a database of firearms featured in movies, TV shows, and video games. \cite{verma2017handheld} constructs a dataset using images from IMFDB as positive samples and randomly collected internet images of flowers, animals, etc., as negative samples. A VGGnet is used to classify and compare the dataset against non-deep learning methods.

In \cite{olmos2018automatic}, a dataset of 9,100 images of people holding handguns is presented. The images are obtained from online gun catalogs and advertisements. They train a VGG network to classify these images for the presence of handguns. To localize the handguns, they use a sliding window approach where each window is sent to the classifier. They also annotated a subset of their dataset with bounding boxes and trained a Faster RCNN for detection.

Later, \cite{qi2021dataset} published a dataset with 51,000 annotated images for gun detection, and most of these images were selected from IMFDB~\cite{imfdb}, and some were from previously published datasets such as \cite{olmos2018automatic}. They fine-tune a Centernet with Mobilenet backbone pretrained on Pascal VOC. Since their final objective was deploying a model onto hardware, they preferred a lightweight model over a more accurate one. 

The imagery data from the above datasets are either non-CCTV or not in the real-world surveillance scene, hence inappropriate for our goal.

\begin{figure*}[]
\begin{center}
\includegraphics[width=0.24\linewidth]{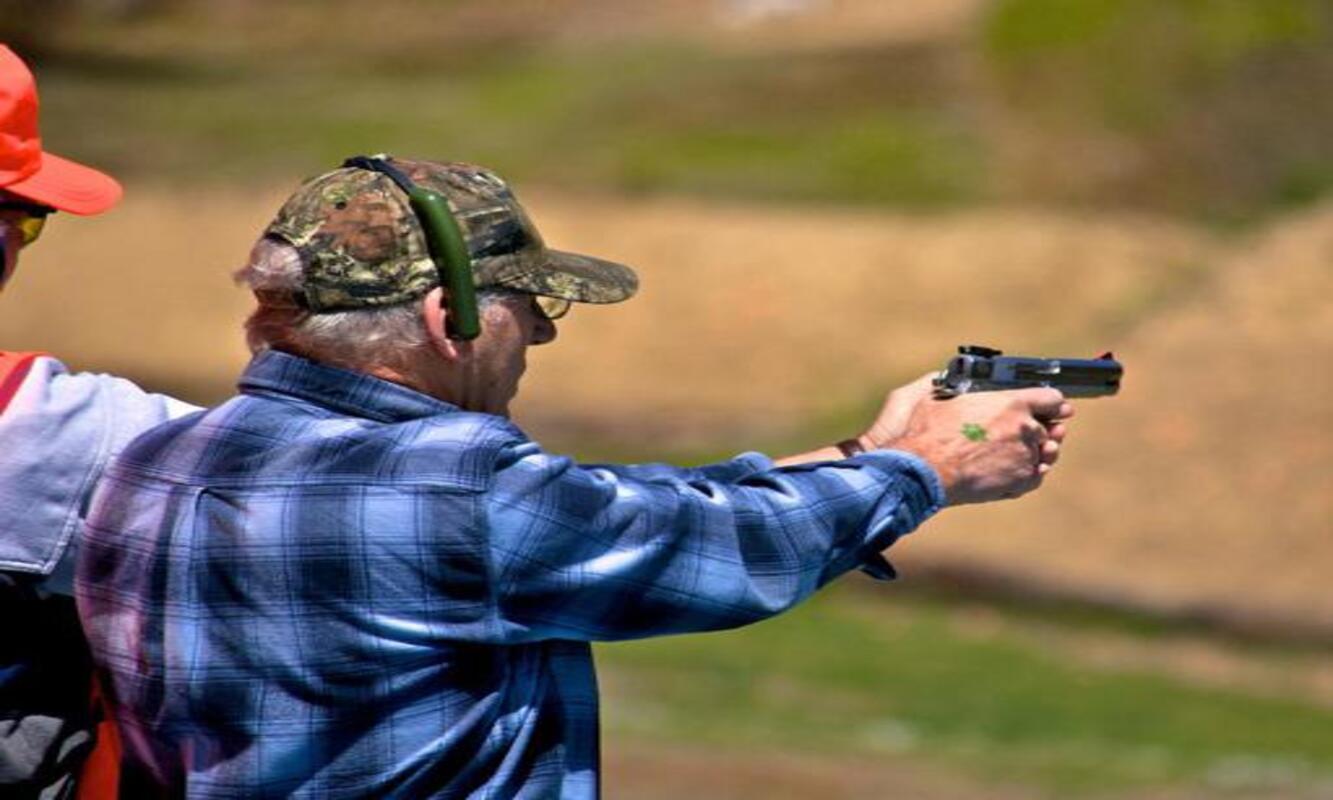} \hfill
\includegraphics[width=0.24\linewidth]{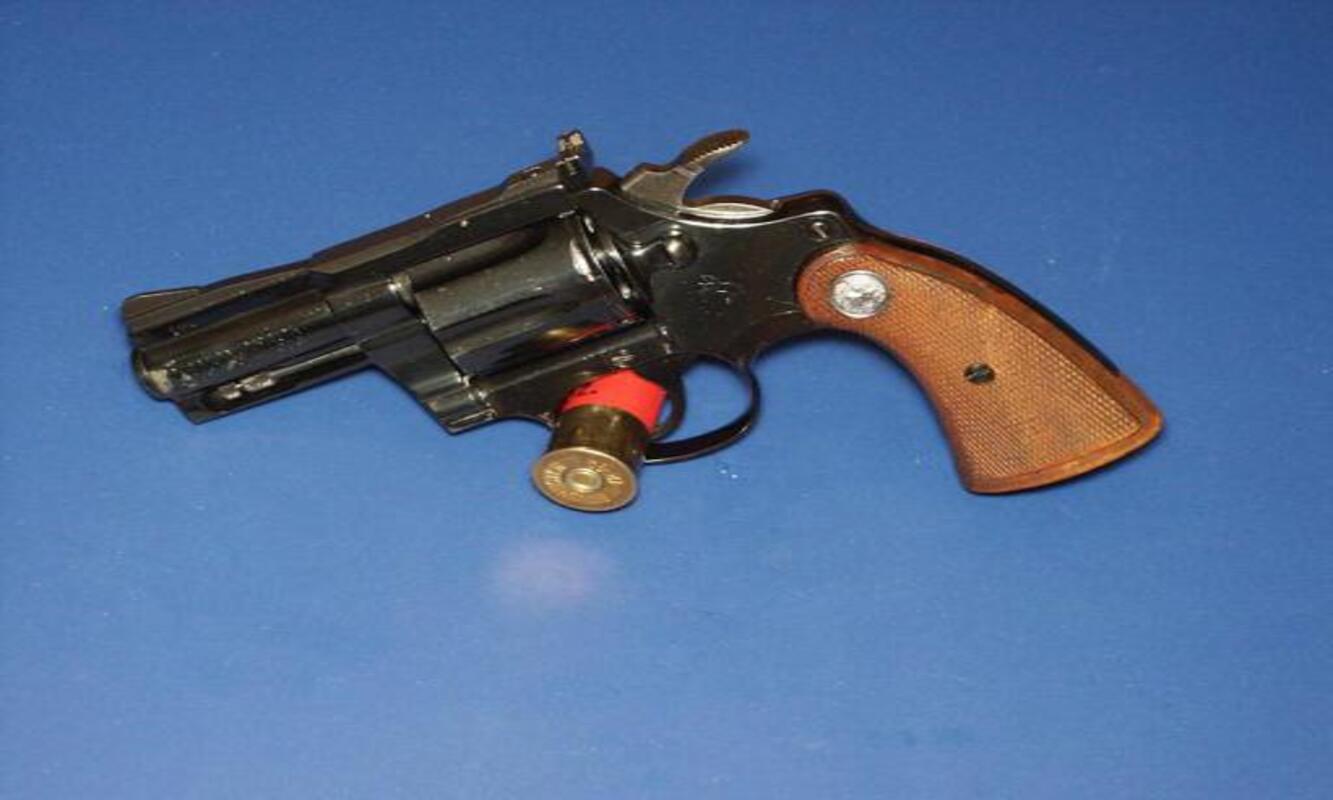} \hfill
\includegraphics[width=0.24\linewidth]{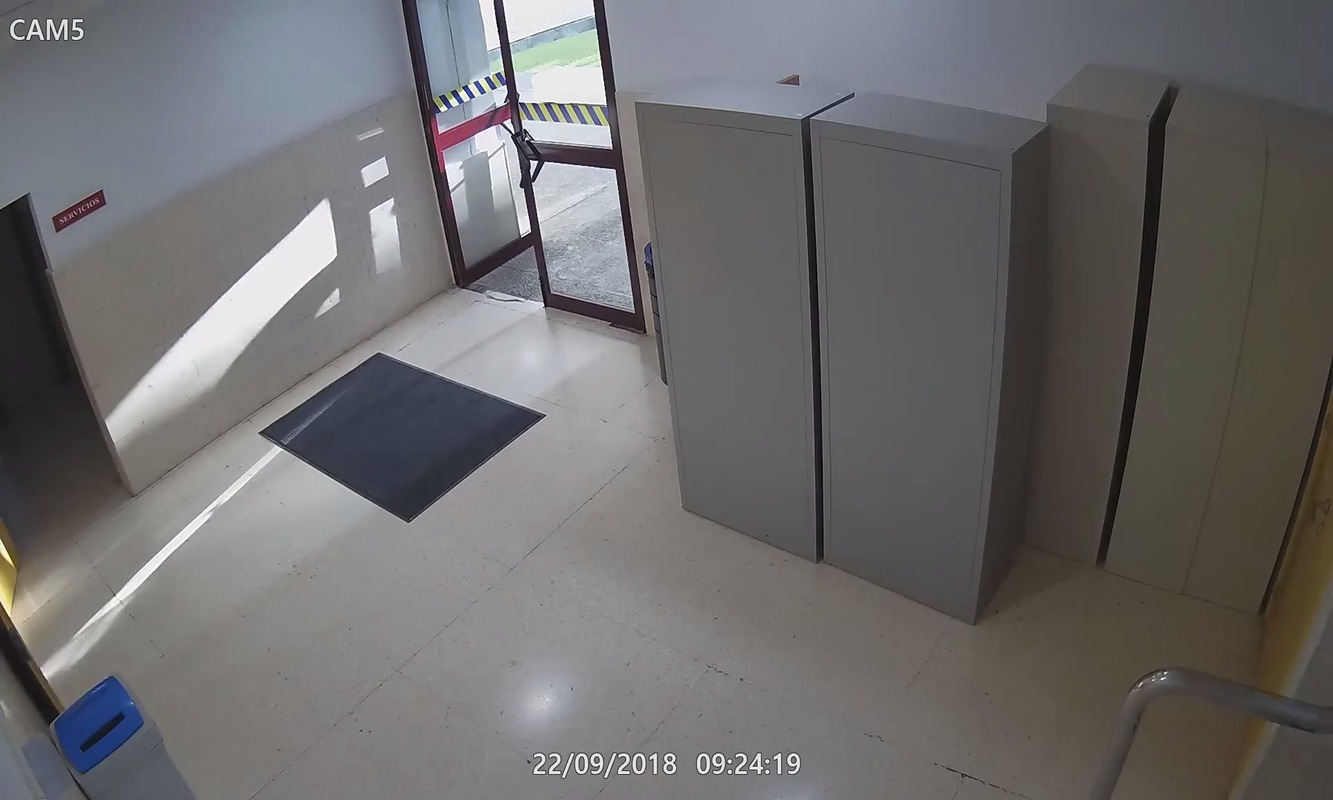} \hfill
\includegraphics[width=0.24\linewidth]{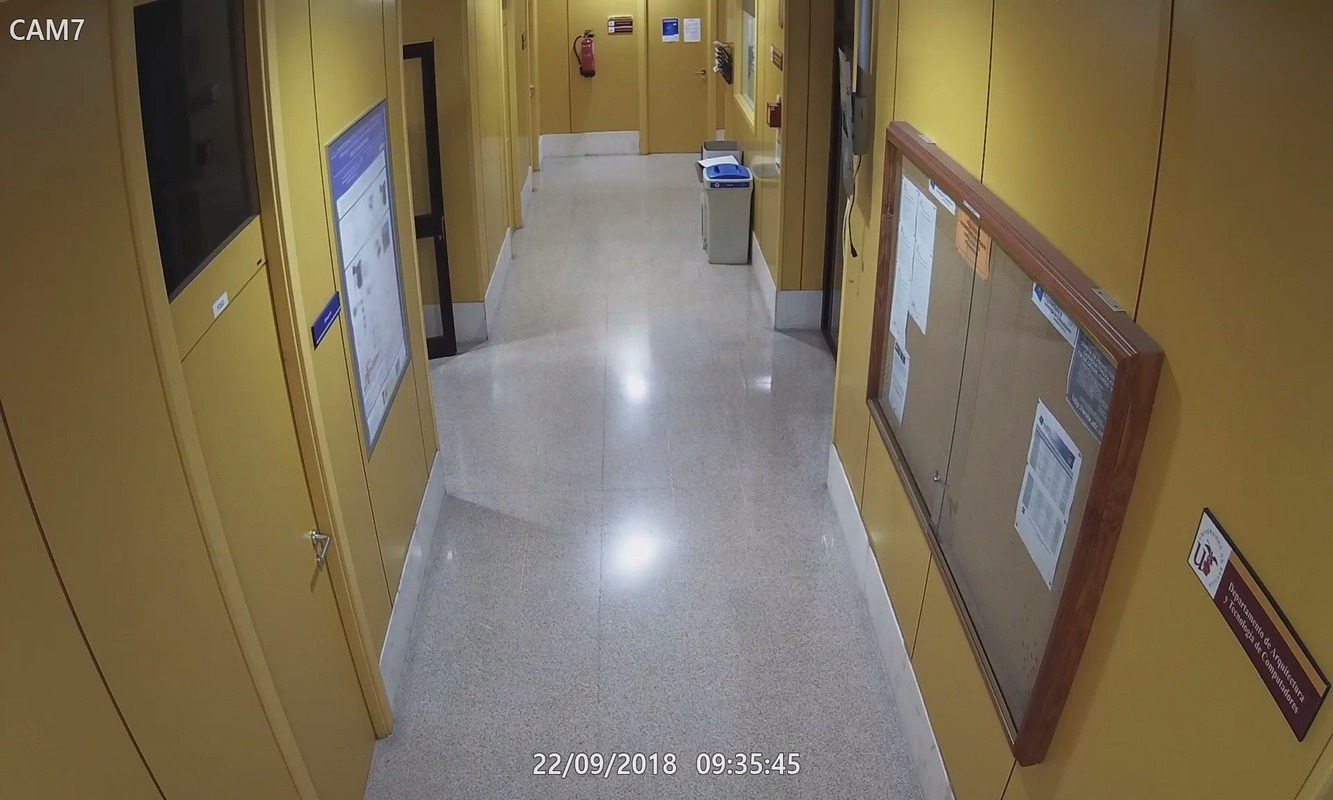}
\end{center}
\vspace{-2mm}  
\caption{Samples of the discarded images. The first two images are examples from the MGD dataset that we discarded since they are stock photos, not from a CCTV perspective. The last two images from the USRT dataset are background images without a person or handgun. }
\label{fig:discarded}
\end{figure*}

\begin{table*}[]
\centering
\begin{tabular}{c|c|c|c|c|c|c}
\hline
Source & Image size & \begin{tabular}[c]{@{}c@{}}\# Unique \\ backgrounds\end{tabular} & \# images & \begin{tabular}[c]{@{}c@{}}\# images\\ with handgun\end{tabular} & \begin{tabular}[c]{@{}c@{}}Avg size of \\ handgun in pixels\end{tabular} & \begin{tabular}[c]{@{}c@{}}Avg size of \\ person in pixels\end{tabular} \\ \hline
MGD \cite{lim2021deep}     & $512 \times 512$    & 41                                                               & 2857      & 2852                                                             & 25                                                                       & 158                                                                     \\ \hline
USRT  \cite{gonzalez2020real}   & $1920\times1080$  & 3                                                                & 3294      & 1115                                                             & 47                                                                       & 319                                                                     \\ \hline
UCF \cite{sultani2018real}    & $320\times240$   & 76                                                               & 1616      & 1597                                                             & 16                                                                       & 79                                                                      \\ \hline
\end{tabular}
\caption{Details of various subsets. Images from the UCF crime scene are much smaller than the other two, making it the most challenging. }
\label{tab:dataset-details}
\end{table*}

In \cite{gonzalez2020real}, the US Real-time Gun detection dataset (USRT) is constructed from a CCTV during a mock attack and annotated for the presence of handguns. USRT hires multiple people holding guns to walk through rooms with CCTV, and 4,118 images are collected. Synthetic data of people with handguns are also generated using the Unity Game engine. They train a Faster RCNN \cite{ren2015faster} network on synthetic data and fine-tune it on the mock attack data.  \cite{lim2021deep} constructs the Monash Gun Dataset (MGD) of 2,500 images, enacting crime scenes recorded with a CCTV. They train an M2Det \cite{zhao2019m2det} model on a pooled dataset of their images and images from \cite{olmos2018automatic}. 

Despite these efforts, detecting guns from real-world CCTV imagery remains underexplored. 
The studies in~\cite{grega2016automated, verma2017handheld} train networks to classify whether an image has a handgun but skip the critical step of gun detection. \cite{olmos2018automatic, qi2021dataset} train Faster R-CNN models to detect handguns, but the images in their datasets are neither from a CCTV perspective nor a real-world surveillance scene. The images in \cite{qi2021dataset} are in ideal conditions and high resolution but mostly are from movie scenes. The dataset constructed by \cite{gonzalez2020real} can be used to pre-train a gun detection model, but they still need to evaluate their model on real crime scene data or analyze SOTA models on the dataset. 

Our work is inspired by the above studies but is the first for thorough benchmarking of handgun detection from real-world CCTV
imagery. On the dataset part, we compile a new benchmark by meticulously selecting appropriate images from USRT and MGD, together with the real-world UCF Crime Scene dataset (UCF)~\cite{sultani2018real}. In addition, we provide richer annotations, enhanced thorough evaluation protocols, and more comprehensive evaluations.

\section{CCTV-Gun Benchmark}

In this section, we describe the details of our CCTV-Gun Benchmark.

\subsection{Dataset Construction}

There has been some preliminary work on handgun detection datasets, but most need improvement. Our dataset, \textbf{CCTV-Gun}, consists of images taken from various CCTV cameras and scenarios. We focus mainly on handguns, the most commonly used type of firearm in gun crimes \cite{zawitz1995guns}. Instead of capturing new images, which is a difficult task,  we seek help from three publicly available datasets: Monash Gun Dataset \cite{lim2021deep}, US Real-time Gun detection dataset \cite{gonzalez2020real}, and UCF Crime scene dataset \cite{sultani2018real}.  More details about images from different subsets can be found in Table \ref{tab:dataset-details}.

The original dataset provided by \cite{lim2021deep} had 7,811 images. Each annotated CCTV image in this dataset contains the presence of a handgun in different outdoor and indoor conditions. The images are of size $512 \times 512$. This dataset had 4,954 stock images obtained from the internet. We discard them as they are not from a CCTV perspective, as seen in Figure \ref{fig:discarded}. We take the remaining 2,857 images from 250 recorded CCTV videos in various indoor and outdoor settings. 

The USRT dataset consists of 5,149 images. These images are from 3 different CCTV cameras, varying lighting conditions, conflicting objects such as fire extinguishers, and often containing multiple people in each frame.  The photos are of size $1920 \times 1080$. This dataset also annotated knives and shotguns, but we ignored them and considered them background. The original dataset provided 5,149 images annotated with handguns. We discard 650 images with no objects, as shown in Figure \ref{fig:discarded}. We take 3294 images from this dataset.  MGD and USRT are mock datasets, meaning the creators have acted out the attack scenes.

UCF Crime scene dataset \cite{sultani2018real} is a large-scale dataset of 128 videos. It contains 1900 untrimmed videos showing 13 anomalies. It is not a gun-detection dataset but a general-purpose anomaly detection dataset. We use the Robbery and Shooting videos (in $320 \times 240$ resolution), which are CCTV camera recordings of real-world crime scenes. We select 57 robbery and 17 shooting videos, extract handgun images in 2 frames/second and obtain 1616 handgun frames.

\subsection{Annotation}

We annotate two objects in each image: the handgun and the person. MGD and USRT datasets already provide handgun annotations, whereas the UCF dataset has no annotations. We provide handgun holder annotations for the first time, making it different from previous works. It is equally important to detect the holder at a potential crime scene. In total, we obtained 7767 annotated images. Among these images, there were 5 images from MGD, 19 from UCF, and 2197 from USRT, which didn't have any handguns. We still include them in the dataset, as they serve as negative examples with a person but no handgun. Examples from our dataset can be seen in Figure \ref{fig:challenge_gt}. We use a graphical image annotation tool \textit{labelimg} \cite{Tzutalin2015} to draw bounding box annotations in our dataset.

We also provide annotations of handgun holder pairs - the person holding each handgun. Although we've not used the pair annotations in training our models, we believe it will benefit the Computer vision community. Using a human-object interaction model, one can refine the handgun features or find the holder for each handgun in an image.

We annotate the test split of our dataset with the following challenges or attributes:
\begin{itemize}
    \vspace{-1.3mm} \item \textbf{Occlusion}: Handguns are often occluded due to the holder's hands at crime scenes, where only a tiny portion of the handgun is visible.
    \vspace{-1.3mm} \item \textbf{Blur}: Since these images are captured from CCTV, some are blurry due to motion.
    \vspace{-1.3mm} \item \textbf{Similar object}: Other small-sized things, such as mobile phones, can be misclassified since handguns occupy a small area in these images.
\end{itemize}

These three attributes were chosen based on visually examining the images. Handguns in MGD images do not have any occlusions, but there are a lot of similar objects, such as mobile phones. In USRT, we found many blurry photos. Since we consider knives and shotguns in USRT as background, they could confuse the detector. We include images with these objects in the ``Similar objects" category. Details can be found in Table \ref{tab:attributes}. Examples of such challenging images are shown in Figure \ref{fig:challenge_gt}.

\setlength{\tabcolsep}{0.5em} 

\begin{table}[]
\begin{tabular}{c|cc|cc|cc}
\hline
\multirow{2}{*}{}                                      & \multicolumn{2}{c|}{Occlusion}  & \multicolumn{2}{c|}{Blur}       & \multicolumn{2}{c}{Similar objects} \\ \cline{2-7} 
                                                       & \multicolumn{1}{c}{USRT} & UCF & \multicolumn{1}{c}{USRT} & MGD & \multicolumn{1}{c}{USRT}    & MGD   \\ \hline
\begin{tabular}[c]{@{}c@{}}\# of\\ images\end{tabular} & \multicolumn{1}{c}{17}   & 34  & \multicolumn{1}{c}{33}   & 17  & \multicolumn{1}{c}{109}     & 29    \\ \hline
\end{tabular}
\caption{Number of images with challenging attributes in each dataset.}

\label{tab:attributes}
\end{table}
\setlength{\tabcolsep}{0.6em}

\subsection{Evaluation Protocols}

We perform two types of experiments: intra-dataset and cross-dataset evaluation.

\subsubsection{Intra-dataset protocol}

\begin{table}[H]
\centering
\begin{tabular}{c|c|c|c|c}
\hline
Source & Total & Train & Val & Test \\ \hline
MGD    & 2857 & 2164  & 287 & 401  \\ \hline
USRT   & 3294 & 2492  & 308 & 494  \\ \hline
UCF    & 1616 & 1435  & 0   & 185  \\ \hline
\end{tabular}
\caption{Train-val-test split of each dataset}
\label{tab:dataset-split}
\end{table}
\subsubsection{Cross-dataset protocol}
Our approach involves training a model on two datasets, $D_1 + D_2$, and testing it on the entirety of $D_3$ without training it on $D_3$. It allows us to assess the generalization ability of SOTA models. To maintain consistency, we use the same training hyperparameters as Intra-dataset, with COCO pretrained models used for initialization.

We then fine-tune the model trained on two datasets on the third dataset. Subsequently, we evaluate the model's performance on the test split of the third dataset. The goal is to compare the benefit of fine-tuning from a model trained on a Gun-detection dataset versus a COCO pretrained model.

\begin{table*}[]
\begin{center}
\begin{tabular}{c|c|c|c|cc|cc|cc}
\hline
\multirow{2}{*}{Backbone} & \multirow{2}{*}{Framework} & \multirow{2}{*}{BS} & \multirow{2}{*}{LR} & \multicolumn{2}{c|}{MGD} & \multicolumn{2}{c|}{USRT} & \multicolumn{2}{c}{UCF} \\ \cline{5-10} 
                          &                            &                     &                     & handgun        & person  & handgun         & person  & handgun        & person \\ \hline
ResNet50                  & Faster RCNN + FPN                & 12                  & 0.01                & 86.8           & 94.8    & 43.7            & 80.0    & 43.4           & 87.2   \\
ResNet50                  & Deformable DETR            & 8                   & 0.0001              & 89.3           & 96.8    & 36.5            & 80.6    & 48.4           & 86.5   \\
ResNet50                  & DetectoRS                  & 4                   & 0.0002              & 87.4           & 95.1    & \textbf{48.9}   & 81.6    & 54.5           & 89.4   \\
Swin-T                    & Faster RCNN + FPN                 & 4                   & 0.01                & \textbf{91.7}  & 93.5    & 44.8            & 86.0    & \textbf{57.4}  & 89.4   \\
ConvNeXt-T                & Faster RCNN + FPN                 & 6                   & 0.0001              & 88.2           & 95.5    & 48.1            & 83.1    & 56.7           & 89.5   \\ \hline
\end{tabular}
\caption{Intra-dataset Performance on different detection models. We train the model on the train split and test on the test split of the same dataset. We compute the average precision at IoU = 0.5 for both handgun and person classes.}
\label{tab:protocol-1}
\end{center}
\end{table*}

\begin{table*}[]
\centering
\begin{tabular}{c|c|cc|cc|cc}
\hline
\multirow{2}{*}{Backbone} & \multirow{2}{*}{Framework} & \multicolumn{2}{c|}{\begin{tabular}[c]{@{}c@{}}Train : MGD + USRT\\ Test : UCF\end{tabular}} & \multicolumn{2}{c|}{\begin{tabular}[c]{@{}c@{}}Train : USRT + UCF\\ Test : MGD\end{tabular}} & \multicolumn{2}{c}{\begin{tabular}[c]{@{}c@{}}Train : UCF + MGD\\ Test : USRT\end{tabular}} \\ \cline{3-8} 
                          &                            & handgun                                       & person                                       & handgun                                       & person                                       & handgun                                       & person                                       \\ \hline
ResNet50                  & Faster RCNN + FPN          & 3.7                                           & 16.0                                           & 47.8                                          & 89.8                                         & 22.1                                          & 86.4                                         \\
ResNet50                  & Deformable DETR            & \textbf{11.7}                                          & 64.4                                         & \textbf{61.2}                                          & 95.8                                         & 15.9                                          & 83.1                                         \\
ResNet50                  & DetectoRS                  & 10.3                                          & 42.0                                           & 60.7                                          & 93.7                                         & 25.7                                          & 88.2                                         \\
Swin-T                    & Faster RCNN + FPN            & 6.8                                           & 17.2                                         & 48.5                                          & 92.1                                         & 26.2                                          & 87.7                                         \\
ConvNext-T                & Faster RCNN + FPN            & 7.7                                           & 32.5                                         & 47.9                                          & 93.0                                           & \textbf{27.3}                                          & 86.4                                         \\ \hline
\end{tabular}
\caption{Cross-dataset performance of models. We compute the average precision at IoU = 0.5}
\label{tab:protocol-2}
\end{table*}

\section{Experiments and Analysis}

\subsection{Setup}

We conduct a thorough evaluation of SOTA object detection algorithms on our dataset. Since we have images and annotations of three different datasets, we perform two kinds of assessment: Intra-dataset and Cross-dataset. We train five object detection methods on our datasets: Faster R-CNN \cite{ren2015faster}, Swin-T \cite{liu2021swin}, Deformable DETR \cite{zhu2020deformable}, DetectoRS \cite{qiao2021detectors}, and ConvNeXt-T \cite{liu2022convnet}. Besides, we use the two-stage with refinement variant of Deformable DETR and employ Cascade RCNN head \cite{cai2018cascade} on DetectoRS. We only use two-stage methods as they are generally more accurate \cite{ren2015faster} and better suited for detecting small objects \cite{redmon2016you}.

 The implementations and COCO pre-trained models are based on MMdetection \cite{chen2019mmdetection}. We train these models for 36 epochs on a 24GB Nvidia A5000 GPU. We decay the learning rate by 0.1 at epochs 27 and 33. Table \ref{tab:protocol-1} provides the training and framework details of the models employed. Our analysis uses the average precision (AP) at IoU = 0.5 for comparing various models. We compute the average precision value for both handgun and person classes. However, \textbf{handgun AP }should be given more weightage, as detecting handguns is a much more challenging task than a person.

\begin{figure*}[]
\begin{center}
\includegraphics[width=0.33\linewidth]{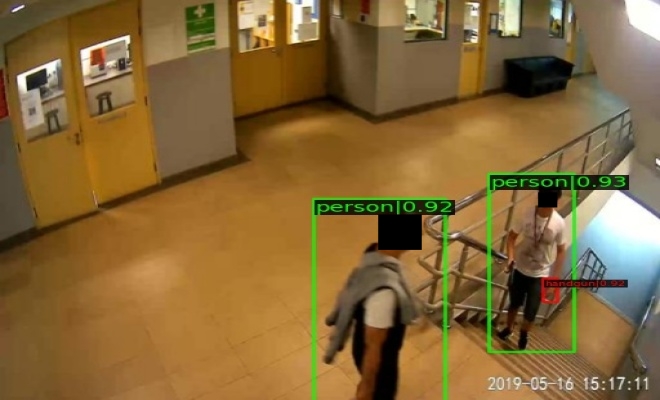} \hfill
\includegraphics[width=0.33\linewidth]{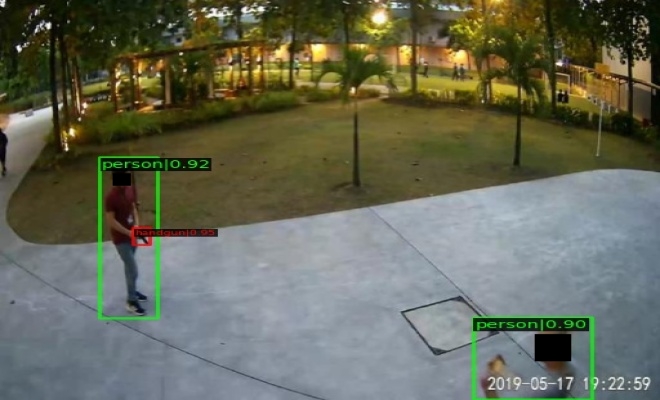} \hfill
\includegraphics[width=0.33\linewidth]{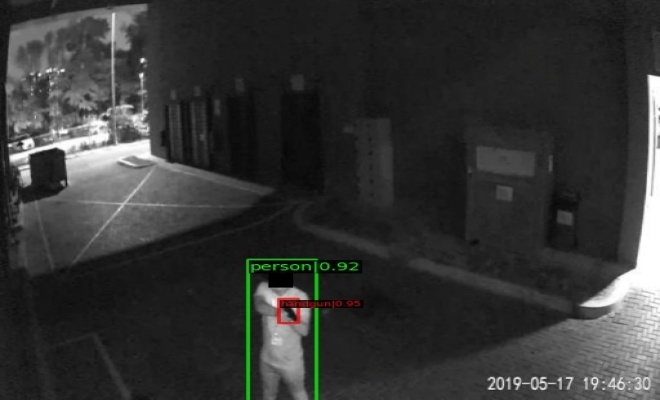} \\
\includegraphics[width=0.33\linewidth]{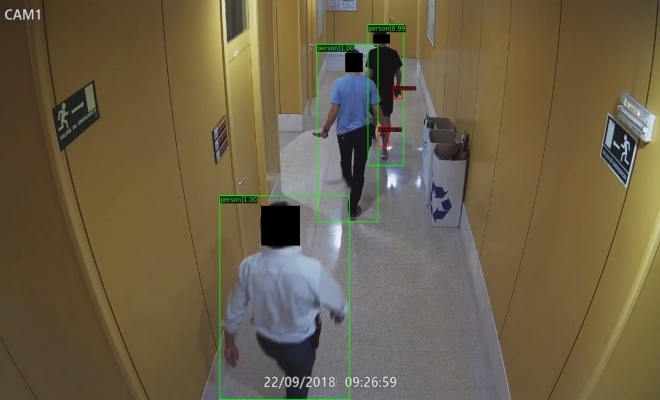} \hfill
\includegraphics[width=0.33\linewidth]{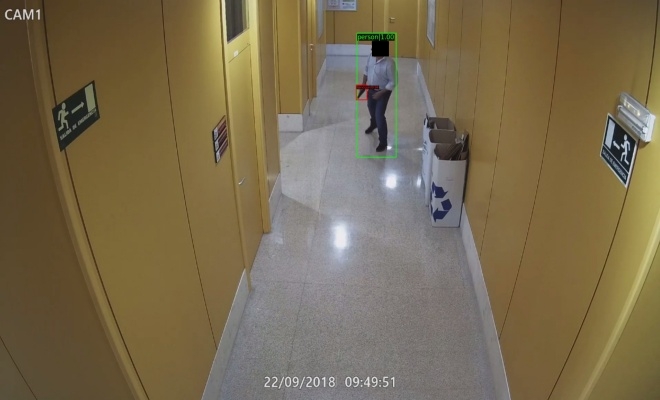} \hfill
\includegraphics[width=0.33\linewidth]{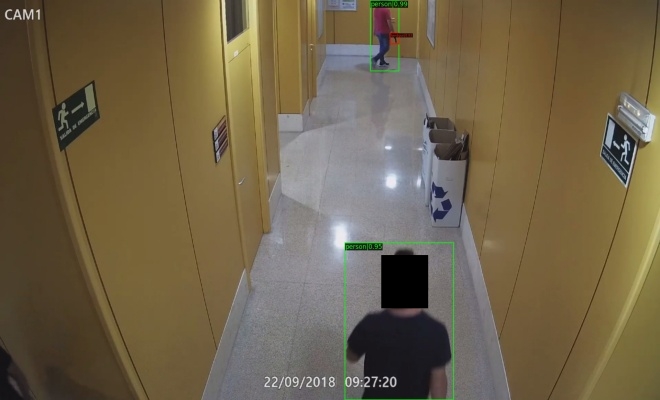} \\
\includegraphics[width=0.33\linewidth]{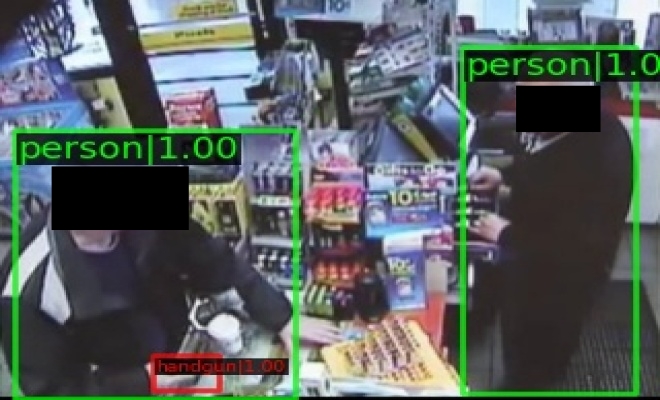} \hfill
\includegraphics[width=0.33\linewidth]{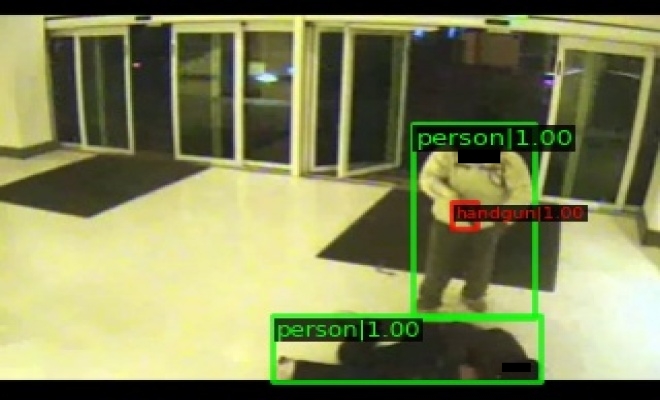} \hfill
\includegraphics[width=0.33\linewidth]{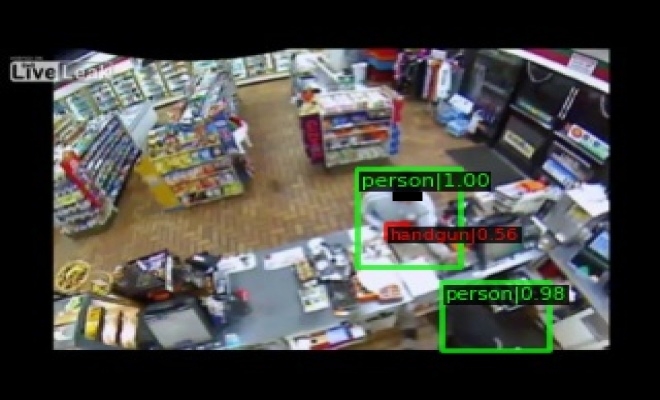} 
\end{center}
\vspace{-2mm}
  \caption{Qualitative results of the best-performing models. Top: MGD, Center: USRT, Bottom: UCF dataset images.}
\label{fig:qualitative_single}
\end{figure*}

\begin{table*}[]
\centering

\begin{tabular}{c|c|cc|cc|cc}
\hline
\multirow{2}{*}{Backbone} & \multirow{2}{*}{Framework} & \multicolumn{2}{c|}{MGD}                                                                                                    & \multicolumn{2}{c|}{USRT}                                                                                                   & \multicolumn{2}{c}{UCF}                                                                                                     \\ \cline{3-8} 
                          &                            & \begin{tabular}[c]{@{}c@{}}COCO\\ pretrained\end{tabular} & \begin{tabular}[c]{@{}c@{}}USRT+UCF\\ pretrained\end{tabular} & \begin{tabular}[c]{@{}c@{}}COCO \\ pretrained\end{tabular} & \begin{tabular}[c]{@{}c@{}}UCF+MGD\\ pretrained\end{tabular} & \begin{tabular}[c]{@{}c@{}}COCO \\ pretrained\end{tabular} & \begin{tabular}[c]{@{}c@{}}MGD+USRT\\ pretrained\end{tabular} \\ \hline
ResNet50                  & Faster RCNN+FPN          & 86.8                                                      & \textbf{87}                                                     & 43.7                                                       & \textbf{47.1}                                                  & \textbf{43.4}                                              & 32.2                                                            \\
ResNet50                  & Deformable DETR            & 89.3                                                      & \textbf{90.5}                                                   & 36.5                                                       & \textbf{39}                                                    & \textbf{48.4}                                              & 43.2                                                            \\
ResNet50                  & DetectoRS                  & 87.4                                                      & \textbf{88.4}                                                   & \textbf{48.9}                                              & 45.9                                                           & \textbf{54.5}                                              & 50.8                                                            \\
Swin-T                    & Faster RCNN+FPN            & 91.7                                                      & \textbf{92.7}                                                   & \textbf{44.8}                                              & 41                                                             & \textbf{57.4}                                              & 47.5                                                            \\
ConvNext-T                & Faster RCNN+FPN            & \textbf{88.2}                                             & 86.8                                                            & 48.1                                                       & \textbf{48.9}                                                  & \textbf{56.7}                                              & 56.7                                                            \\ \hline
\end{tabular}
\caption{Effectiveness of pre-training on handgun datasets. In COCO pretrained column, models were pre-trained on COCO and then fine-tuned on target. In USRT + UCF pretrained column, models were pretrained on COCO, then USRT + UCF, and finally fine-tuned on the target dataset. We compute the handgun AP score at IoU = 0.5.}
\label{tab:protocol_3}
\end{table*}

\subsection{Results and analysis}

\subsubsection{Intra dataset protocol}
Table \ref{tab:protocol-1} presents the results of Intra-dataset evaluation. All five models perform well on the MGD dataset, with Swin-T achieving the highest handgun AP score. Swin-T and DetectoRS were the top-performing models on UCF and USRT datasets. Figure \ref{fig:qualitative_single} provides qualitative results. It is noteworthy that despite MGD and USRT being high-resolution images, the models performed considerably worse on the USRT dataset. There are several possible reasons to explain this behavior:

Firstly, the USRT dataset has fewer positive examples of handguns, with only 1115 out of 3294 images featuring handguns. In contrast, handguns are present in almost all the images from the MGD dataset, providing detection models with fewer positive examples. Secondly, several images from USRT include similar objects like knives and torch lights. This makes the detection task much more challenging, as the model is prone to confusion between handguns and these other objects. Finally, we ignored larger guns like Shotguns and Assault rifles in the annotation, treating them as background, which makes it more difficult.

\begin{figure*}
     \centering

         \begin{subfigure}[b]{0.33\textwidth}
         \centering
         \includegraphics[width=\textwidth]{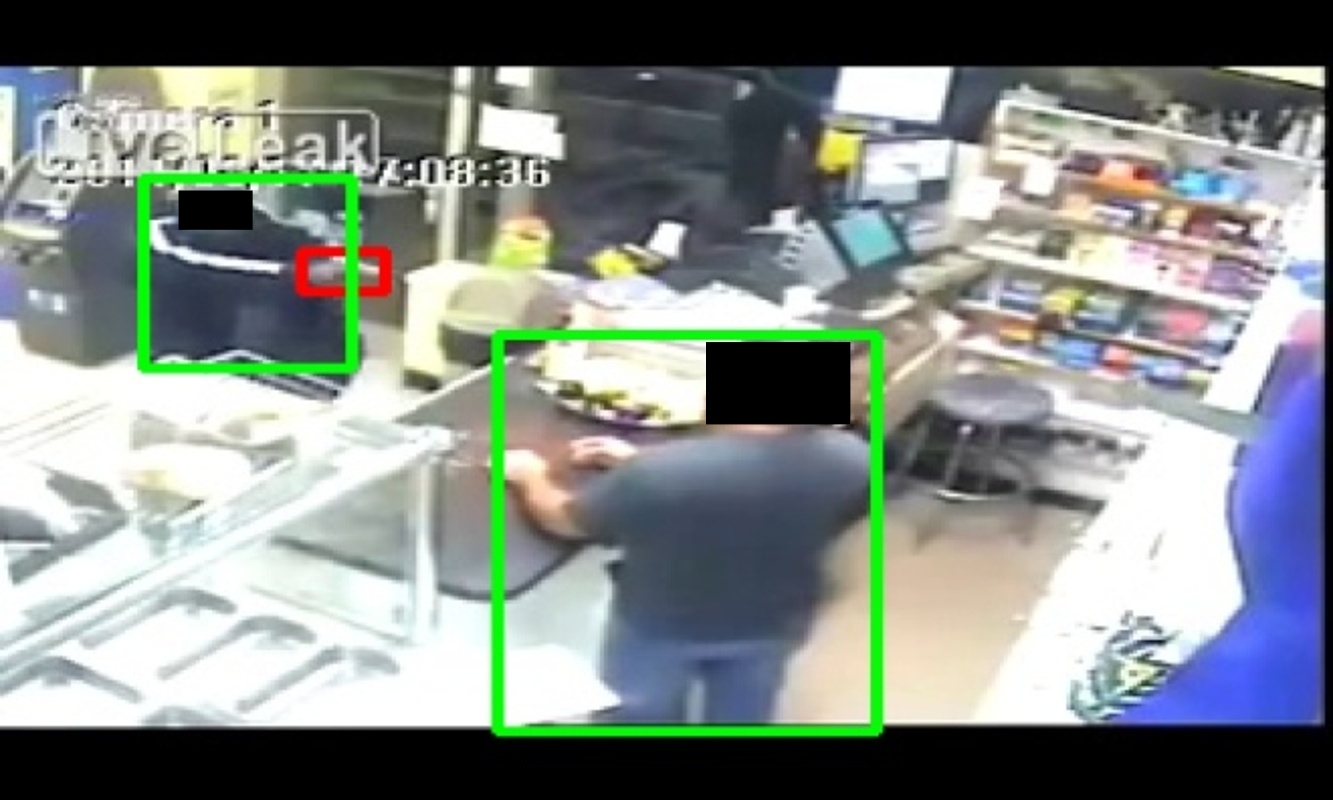}
         
     \end{subfigure}
     \hfill
     \begin{subfigure}[b]{0.33\textwidth}
         \centering
         \includegraphics[width=\textwidth]{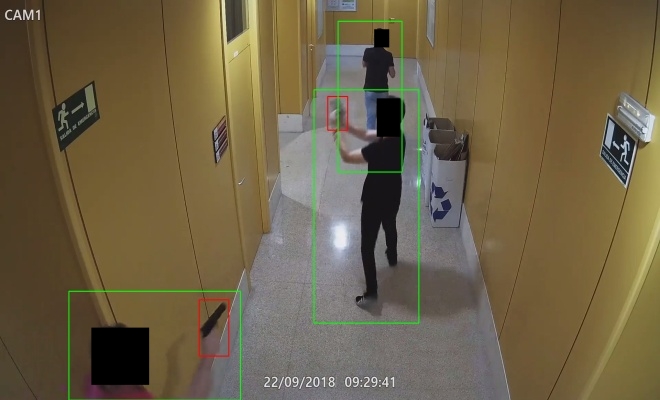}
         
     \end{subfigure}
     \hfill
     \begin{subfigure}[b]{0.33\textwidth}
         \centering
         \includegraphics[width=\textwidth]{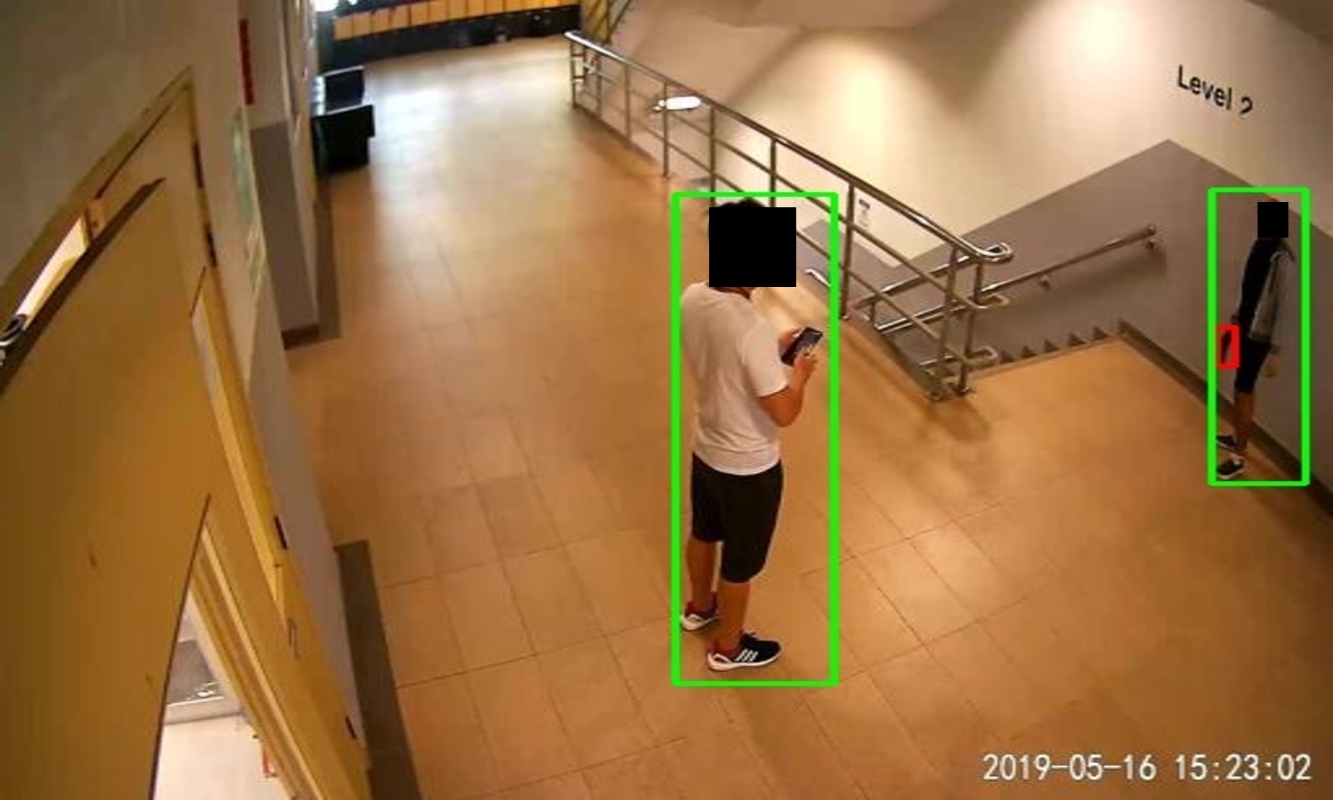}
         
     \end{subfigure}\\
\vspace{-4mm}     \begin{subfigure}[b]{0.33\textwidth}
         \centering
         \includegraphics[width=\textwidth]{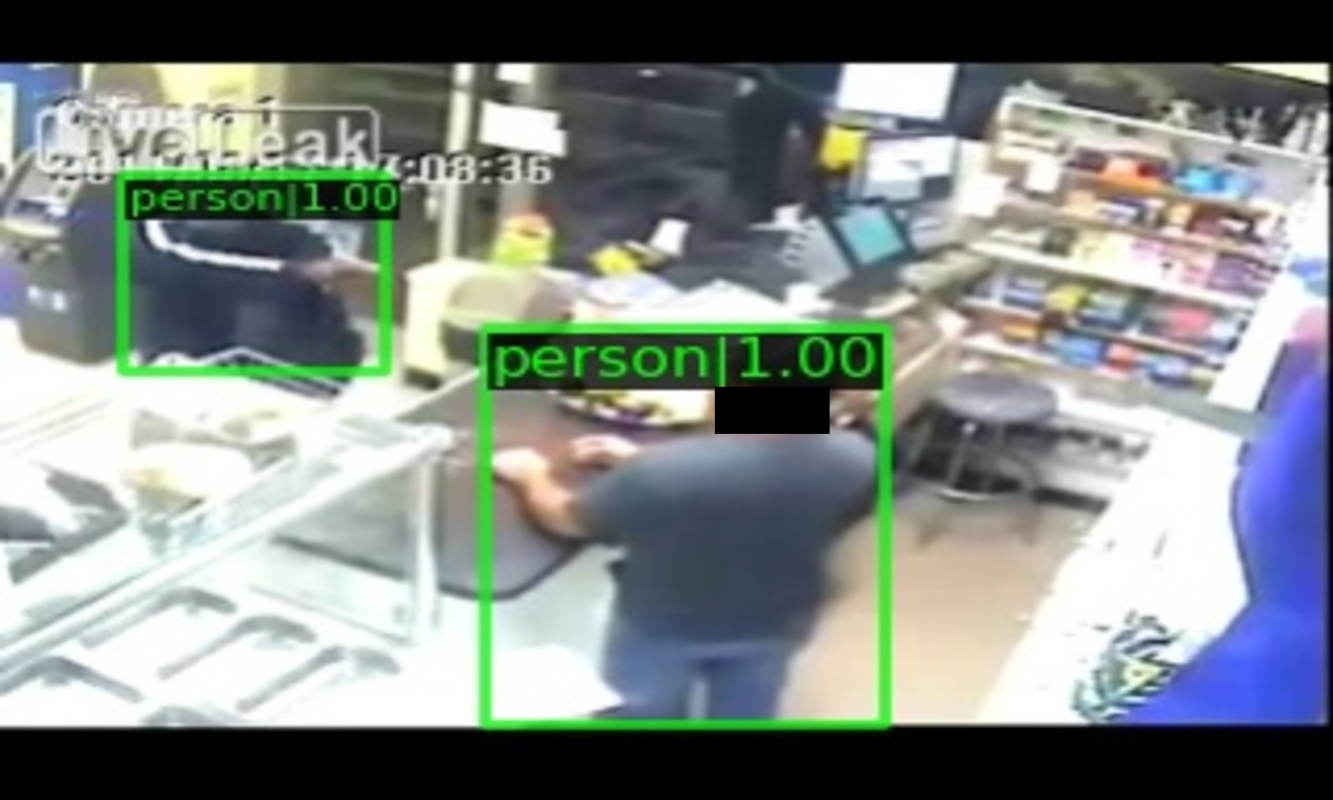}
         \caption{Occlusion}
         
     \end{subfigure}
     \hfill
     \begin{subfigure}[b]{0.33\textwidth}
         \centering
         \includegraphics[width=\textwidth]{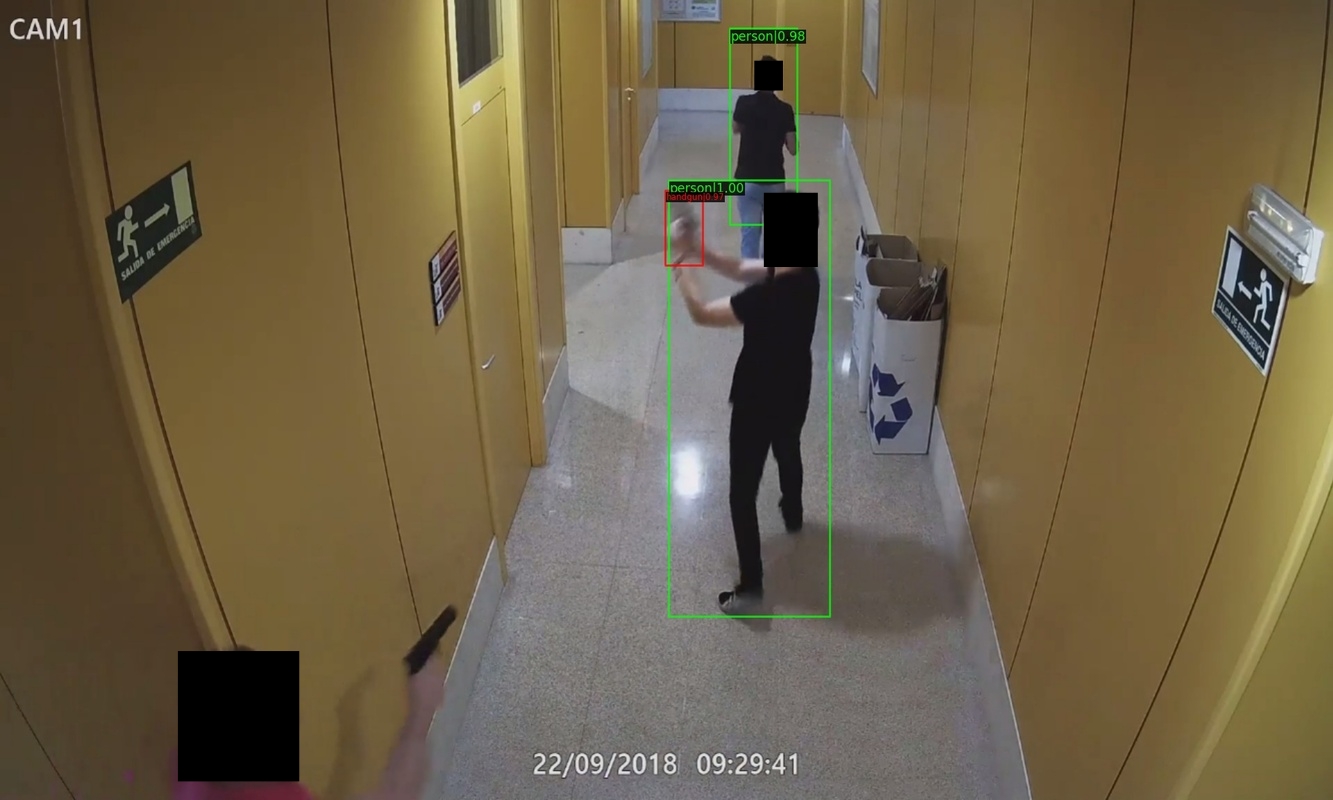}
         \caption{Blur}
         
     \end{subfigure}
     \hfill
     \begin{subfigure}[b]{0.33\textwidth}
         \centering
         \includegraphics[width=\textwidth]{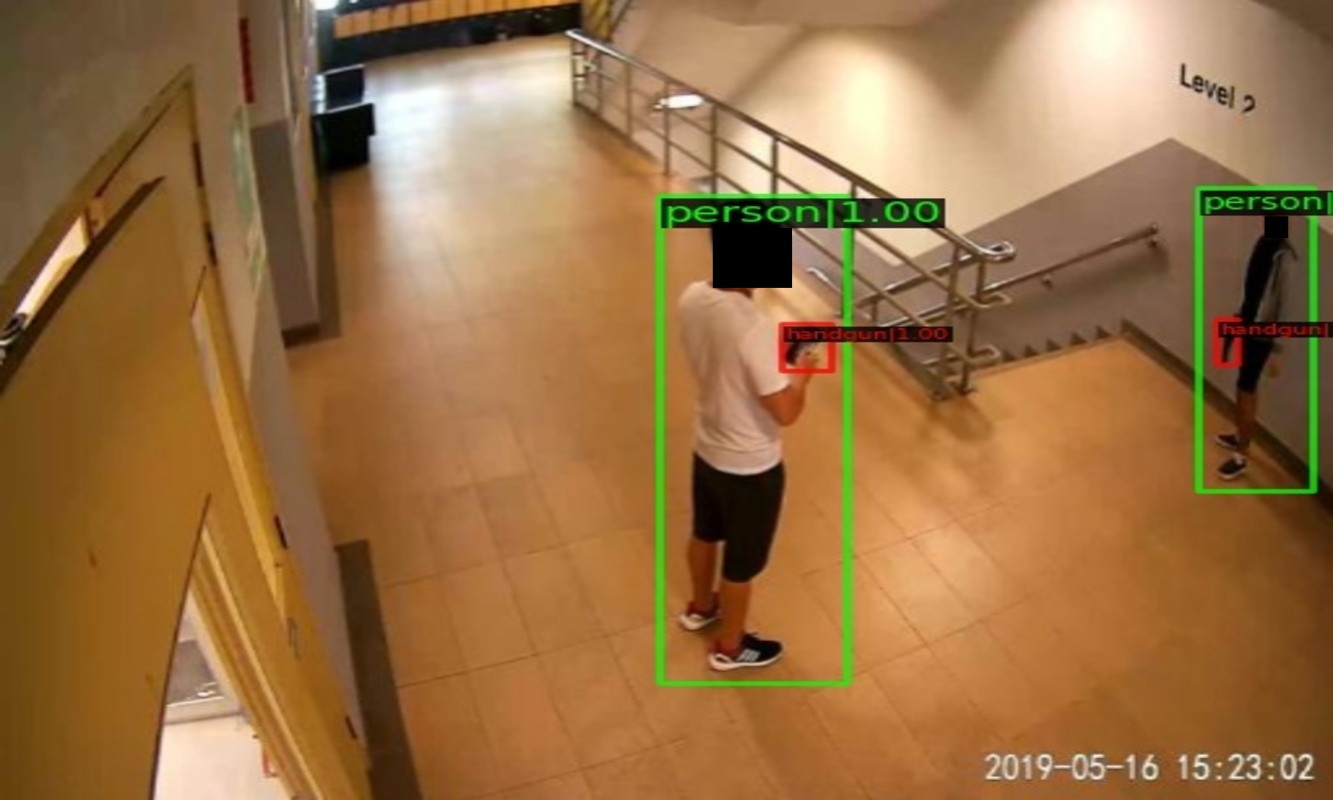}
         \caption{similar objects}
         
     \end{subfigure}
\caption{Misclassifications from the best performing models on the challenging tasks. The First row contains the ground truth, and the second shows the network output.}
        \label{fig:challenge_misclass}
\end{figure*}

\begin{table*}[]
\centering
\begin{tabular}{c|c|cc|cc|cc}
\hline
\multirow{2}{*}{Backbone} & \multirow{2}{*}{Framework} & \multicolumn{2}{c|}{Occlusion}  & \multicolumn{2}{c|}{Blur}       & \multicolumn{2}{c}{Similar objects} \\ \cline{3-8} 
                          &                            & USRT           & UCF            & MGD            & USRT           & MGD               & USRT             \\ \hline
ResNet50                  & Faster RCNN + FPN          & 50.92          & 38.92          & 76.63          & 34.36          & 69.50             & 45.05            \\
ResNet50                  & Deformable DETR            & 16.03          & 20.64          & \textbf{77.46} & 23.42          & 70.15             & 22.08            \\
ResNet50                  & DetectoRS                  & 25.97          & 27.74          & 77.28          & 29.91          & 67.81             & 45.03            \\
Swin-T                    & Faster RCNN + FPN            & 45.56          & 23.34          & 76.40          & 40.10          & \textbf{73.81}    & 47.74            \\
ConvNext-T                & Faster RCNN + FPN            & \textbf{62.76} & \textbf{44.14} & 75.52          & \textbf{42.83} & 69.87             & \textbf{49.56}   \\ \hline
\end{tabular}
\caption{Performance of detection models on challenging attributes. We report the average precision of the handgun class computed on the images with selected attributes.}
\label{tab:challenge_perf}
\end{table*}

\subsubsection{Cross dataset protocol}
Results for Cross-dataset evaluation (without fine-tuning) can be found in Table \ref{tab:protocol-2}. We observe that models trained on MGD + USRT perform poorly on the UCF dataset. MGD and USRT are made of enacted crime scenes with clear, high-resolution frames, whereas UCF data comprises real crime scene images taken at low resolution. Models trained on USRT + UCF perform pretty well on MGD since images in the MGD dataset are clear images with very few occlusions. This trend is also seen in the Intra-dataset setting, where models achieve high handgun AP scores on MGD's test set. 

Fine-tuning results can be found in Table \ref{tab:protocol_3}. Only when fine-tuned on MGD did the Gun-detection (USRT+UCF) trained model performs better than COCO pretrained. The effectiveness is inconclusive for USRT, where the Gun-detection (UCF+MGD) trained model performs better in 2 out of 5 cases. In UCF, we observed worse performance when the MGD+USRT model was used for fine-tuning. Since MGD is the least challenging dataset among the three, pre-training may have helped. UCF, with its small images and heavy occlusions, obtains more significant benefits when a COCO pretrained model is used for fine-tuning. The combined size of MGD + USRT might not have been enough to act as an effective pre-training dataset in this case.

\subsection{Challenging attributes}
We annotate the test split of our dataset with three challenging attributes: occlusions, blur, and similar objects. We then evaluate models trained on each dataset on these attributes. We report the average handgun precision at IoU = 0.5 for these models in Table \ref{tab:challenge_perf} and provide failure cases in Figure \ref{fig:challenge_misclass}. Results are similar to Intra-dataset evaluation - models which perform well are more robust towards challenges. ConvNeXt-T achieves the best handgun AP score in 4 out of 6 categories. 

\subsection{Discussion}

The results presented in the experiments section demonstrate the effectiveness of the proposed handgun detection benchmark in evaluating different models' performance. The evaluation was carried out on Intra-datasets, Cross-dataset settings, and challenging attributes, providing a comprehensive assessment of the model's strengths and limitations.

The Intra-dataset evaluation showed that the performance of the models varied significantly across different datasets. The models performed best on MGD, which had clear, high-resolution frames. They struggled with USRT, which had fewer positive examples of handguns. The Cross-dataset evaluation without fine-tuning revealed that models trained on MGD + USRT performed poorly on UCF, with low-resolution images taken at real crime scenes. However, models trained on USRT + UCF performed well on MGD, which had clear images with very few occlusions. Fine-tuning the models on different datasets showed mixed results, suggesting that pre-training on specific datasets may or may not be effective depending on the target dataset's characteristics.

Finally, evaluating challenging attributes demonstrated that the models performing well on individual datasets were more robust against occlusion, blur and similar objects. ConvNeXt-T achieved the best results in most categories, highlighting its robustness towards challenging scenarios.

\section{Conclusion}
In conclusion, gun-related violence is a significant security issue in many countries, especially the United States. Detecting handguns in real-world CCTV imagery can potentially prevent gun-related crimes and enhance public safety. However, the small size of handguns in crime CCTV images, occlusions caused by the holder's hands, and varying lighting conditions make handgun detection challenging. To address these challenges, we presented \textbf{ CCTV-Gun}, a meticulously crafted and annotated benchmark for real-world handgun detection from CCTV images. Through detailed bounding box annotations for persons, handguns, and handgun holder pairs, combined with the evaluation protocol and thorough experiments, our benchmark provides a valuable resource for training and evaluating handgun detection algorithms. We hope that the availability of this benchmark will facilitate further research in this area and encourage the development of more effective solutions to address the serious issue of gun violence.

\paragraph{Acknowledgement.} We thank Sabbarish Ramana Rajan for their contribution to the early study of the work. We also thank the authors of the  UCF Crime Scene dataset (UCF) \cite{sultani2018real}, US Real-time Gun detection dataset (USRT) \cite{gonzalez2020real}, and Monash Gun Dataset (MGD) \cite{lim2021deep} for providing the datasets that form the base of our work. 

{\small
\bibliographystyle{ieee_fullname}
\bibliography{references}
}

\end{document}